\documentclass[runningheads]{llncs}
\usepackage{graphicx}
\usepackage{comment}
\usepackage{amsmath,amssymb} 
\usepackage{color}

\usepackage{algorithm}
\usepackage{algorithmicx}
\usepackage{algpseudocode} 
\usepackage{multirow}
\usepackage{multicol}
\usepackage{wrapfig}

\newcommand{\vectorize}[1]{{\rm vec}(#1)}

\begin{document}
\pagestyle{headings}
\mainmatter
\def\ECCVSubNumber{2810}  

\title{Differentiable Feature Aggregation Search for Knowledge Distillation} 

\titlerunning{Differentiable Feature Aggregation Search for Knowledge Distillation}
%
\author{Yushuo Guan$^{1}$\thanks{These authors contributed equally to this work.} \and
Pengyu Zhao$^{1\star}$ \and
Bingxuan Wang$^{1}$ \and
Yuanxing Zhang$^{1}$ \and
Cong Yao$^{2}$ \and
Kaigui Bian$^{1,3}$ \and
Jian Tang$^{4}$ \\
$^{1}$Peking University, $^{2}$Megvii (Face++) Technology Inc, $^{3}$National Engineering Laboratory for Big Data Analysis and Applications, $^{4}$DiDi AI Labs \\
Emails: \email{\{david.guan,pengyuzhao,wangbx,longo\}@pku.edu.cn}, \email{yaocong2010@gmail.com}, \email{bkg@pku.edu.cn}, \email{tangjian@didiglobal.com}
}
\authorrunning{Y. Guan et al.}
%
\institute{}
\maketitle

\begin{abstract}
Knowledge distillation has become increasingly important in model compression. It boosts the performance of a miniaturized student network with the supervision of the output distribution and feature maps from a sophisticated teacher network. Some recent works introduce multi-teacher distillation to provide more supervision to the student network. However, the effectiveness of multi-teacher distillation methods are accompanied by costly computation resources. 
To tackle with both the efficiency and the effectiveness of knowledge distillation, we introduce the feature aggregation to imitate the multi-teacher distillation in the single-teacher distillation framework by extracting informative supervision from multiple teacher feature maps. Specifically, we introduce DFA, a two-stage Differentiable Feature Aggregation search method that motivated by DARTS in neural architecture search, to efficiently find the aggregations.
In the first stage, DFA formulates the searching problem as a bi-level optimization and leverages a novel bridge loss, which consists of a student-to-teacher path and a teacher-to-student path, to find appropriate feature aggregations. The two paths act as two players against each other, trying to optimize the unified architecture parameters to the opposite directions while guaranteeing both expressivity and learnability of the feature aggregation simultaneously. In the second stage, DFA performs knowledge distillation with the derived feature aggregation.
Experimental results show that DFA outperforms existing distillation methods on CIFAR-100 and CINIC-10 datasets under various teacher-student settings, verifying the effectiveness and robustness of the design.
\end{abstract}

\section{Introduction}
In recent years, visual recognition tasks have been significantly improved by deeper and larger convolutional networks. However, it is difficult to directly deploy such complicated networks on certain computationally limited platforms such as robotics, self-driving vehicles and most of the mobile devices. Therefore, the community has raised increasing attention on model compression approaches such as model pruning~\cite{dong2019network,li2016pruning,wen2016learning}, model quantization~\cite{hubara2016binarized,leng2018extremely,lin2017towards} and \textit{knowledge distillation}~\cite{hinton2015distilling,romero2014fitnets,tung2019similarity,you2017learning,zagoruyko2016paying}.  

\begin{figure}[htbp]
\centering
\includegraphics[width=\linewidth]{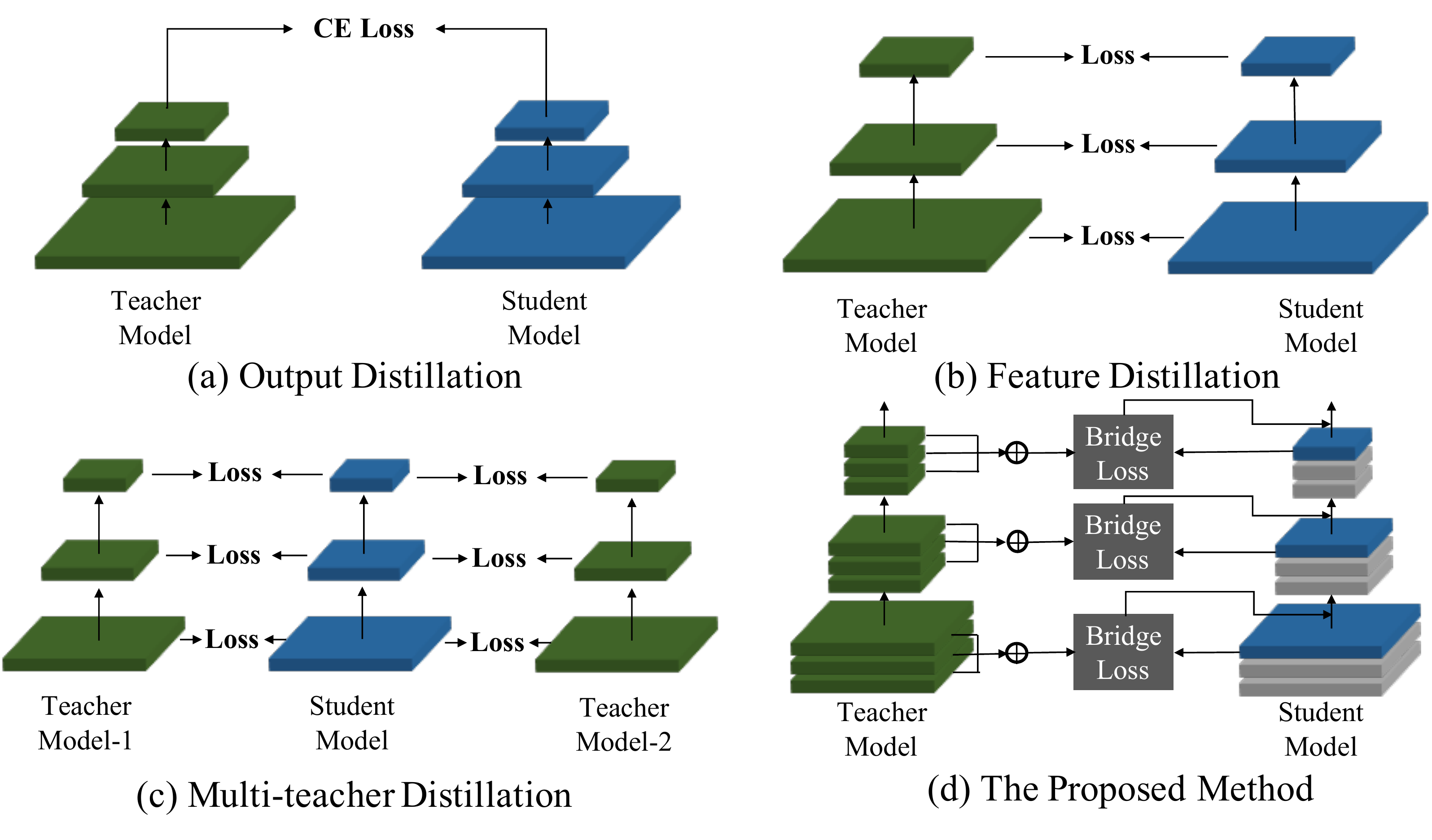}
\caption{Illustrations of different knowledge distillation methods. (a) Output distillation. (b) Feature distillation. (c) Multi-teacher distillation. (d) DFA leverages a novel bridge loss for feature distillation, which takes the advantage of NAS and feature aggregation.}
\label{fig:overview}
\end{figure}

Knowledge distillation refers to the methods that supervise the training of a small network (\textit{student}) by using the knowledge extracted from one or more well-trained large networks (\textit{teacher}). 
The key idea of knowledge distillation is to transfer the knowledge from the teacher networks to the student network. The first attempt of the knowledge distillation for deep neural networks leverages both the correct class labels and the soft targets of the teacher network, i.e., the soft probability distribution over classes, to supervise the training of the student network.
The recent advances of knowledge distillation can be mainly divided into two categories: \textit{output distillation}~\cite{hinton2015distilling} and \textit{feature distillation}~\cite{romero2014fitnets,tung2019similarity,zagoruyko2016paying}, as shown in Fig.~\ref{fig:overview} (a-b).
More recent works concentrate on multi-teacher distillation with \textit{feature aggregation}~\cite{you2017learning}, where an ensemble of teacher networks provide richer information from the aggregation of output distributions and feature maps. Although an ensemble of teacher networks could
provide richer information from the aggregation of output
distributions and feature maps, they require much more computation
resources than single-teacher distillation. 

To achieve the same effect as the multi-teacher distillation with less computation overheads, we propose DFA, a two-stage Differentiable Feature Aggregation search method in the single-teacher knowledge distillation by coupling features from different layers of a single network as multiple ``teachers'', and thus avoids the computation expenses on running several large teacher networks.
Specifically, DFA first searches for the appropriate feature aggregation, i.e., the weighted sum of the feature maps, for each layer group in the teacher network by finding the best aggregation weights.
Then, it conducts the normal feature distillation with the derived aggregations.
Inspired by DARTS~\cite{liu2019darts}, DFA leverages the differentiable group-wise search in the first stage, which formulates the searching process as a bi-level optimization problem with feature aggregation weights as the upper-level variable and the model parameters as the lower-level variable. 
Moreover, as the common distillation loss and cross-entropy loss fail to find the appropriate feature aggregations, a novel \textit{bridge loss} is introduced as the objective function in DFA, where (1) a student-to-teacher path is built for searching the layers that match the learning ability of student network, and (2) a teacher-to-student path is established for finding the feature aggregation with rich features and a wealth of knowledge.
Experiments on CIFAR-100~\cite{krizhevsky2009learning} and CINIC~\cite{darlow2018cinic} datasets show that DFA could outperform the state-of-the-art distillation methods, demonstrating the effectiveness of the feature aggregation search.
%

The main contributions of this paper are as follows:
\begin{itemize}
    \item We introduce DFA, a Differentiable Feature Aggregation search method to mimic multi-teacher distillation in the single-teacher distillation framework, which first searches for appropriate feature aggregation weights and then conducts the distillation with the derived feature aggregations.
    \item We propose a novel bridge loss for DFA. The bridge loss consists of a student-to-teacher path and a teacher-to-student path, which simultaneously considers the expressivity and learnability for the feature aggregation.
    \item Experimental results show that the performance of DFA surpasses the feature aggregations derived by both hand-crafted settings and random search, verifying the strength of the proposed method.
\end{itemize}

\section{Related Work}

\noindent \textbf{Knowledge Distillation}: Knowledge distillation~\cite{bucilu2006model} is firstly introduced in model compression. 
Despite the classification loss, the student network is optimized by an extra cross-entropy loss with the soft target from the teacher network, i.e. the probability distribution softened by temperature scaling. 
Hinton et al.~\cite{hinton2015distilling} employ knowledge distillation in the training of deep neural networks. However, with the huge gap of model capacity among the neural networks, it is hard for the student to learn from the output distribution of a cumbersome teacher directly. 
Thus, several approaches~\cite{kim2018paraphrasing,romero2014fitnets,zagoruyko2016paying} exploit feature distillation in the student training, where the student network mimics the feature maps from the teacher network of different layers. Multi-teacher knowledge distillation~\cite{you2017learning} takes a further step, which takes full advantage of the feature maps and the class distributions amalgamated from an ensemble of teacher networks. The \textit{feature aggregation} from multiple teachers helps the student learn from different perspectives. However, compared with single-teacher distillation, more computation resources are demanded for extracting useful information from all the teachers.

\noindent \textbf{Neural Architecture Search}:
With the vigorous development of deep learning, neural architecture search (NAS), an automatic method for designing the structure of neural networks, has been attracting increasing attention recently.
The early works mainly sample and evaluate a large number of networks from the search space, and then train the sampled models with reinforcement learning~\cite{cai2018path,tan2019mnasnet,zoph2017neural,zoph2018learning} or update the population with the evolutionary algorithm~\cite{real2019regularized,real2017large}.
Though achieving state-of-the-art performance, the above works are all computation expensive. 
Recent works propose the one-shot approaches~\cite{pmlr-v80-bender18a,cai2019proxylessnas,dong2019one,liu2019darts,pham2018efficient,xie2019snas} in NAS to reduce the computation cost.
It models NAS as a single training process for an over-parameterized network covering all candidate sub-networks named supernet, and then selects the network architecture from the trained supernet.
Among the one-shot methods, the differentiable architecture search (DARTS)~\cite{chen2019progressive,li2020neural,liu2019darts,nayman2019xnas,xu2020pc,zela2020understanding} further relaxes the discrete search space to be continuous and couples the architecture parameters with the model parameters in the supernet. 
Therefore, the architecture parameters can be jointly optimized in the one-shot training along with the model parameters by gradient descent.

There have been several methods designed for combining NAS with knowledge distillation.
DNA~\cite{li2019blockwisely} searches for the light-weight architecture of the student network from a supernet. KDAS~\cite{kang2019towards} builds up the student network progressively based on an ensemble of independently learned student networks.
As opposed to these methods, we try to imitate the multi-teacher distillation in the single teacher distillation framework by finding the appropriate feature aggregations in the teacher network with differentiable search strategy.


\begin{figure*}[t]
\centering
\includegraphics[width=0.8\textwidth]{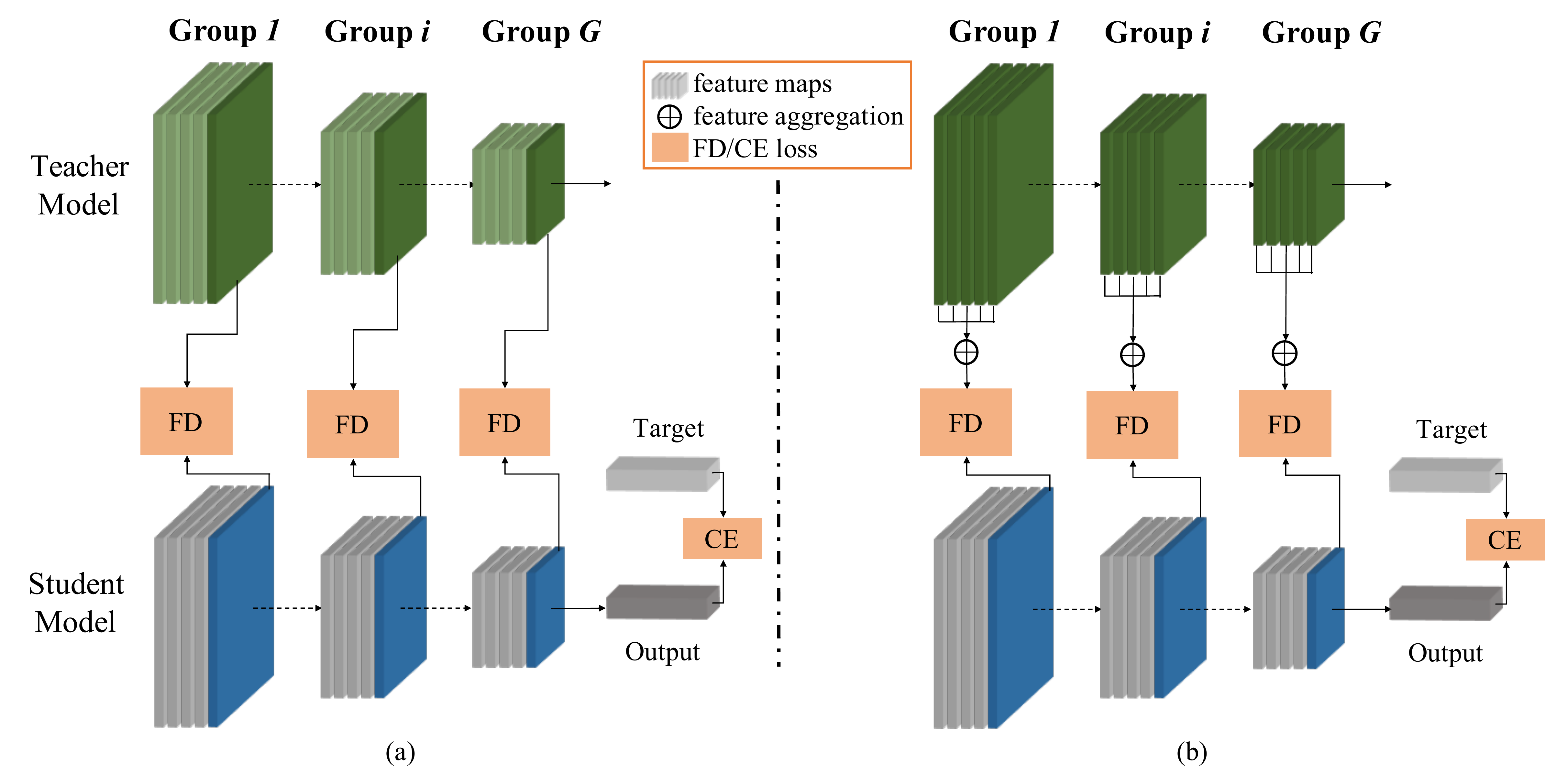}
\caption{Comparisons between traditional feature distillation and DFA. (a) Traditional methods implement distillation with the last feature map in each layer group of the teacher network. (b) DFA leverages the feature aggregation of the teacher for distillation, which contains rich features and a wealth of knowledge. ``FD'' and ``CE'' represent the feature distillation loss $\mathcal{L}_{\text{fd}}$ and cross entropy loss $\mathcal{L}_{\text{ce}}$ respectively.}
\label{fig:compare}
\end{figure*}

\section{Method}
We propose the two-stage Differentiable Feature aggregation (DFA) method for single-teacher knowledge distillation, as outlined in Algorithm~\ref{alg}. In the first stage (i.e., ``AGGREGATION SEARCH''), DFA searches for appropriate feature aggregation weights. In the second stage (i.e., ``FEATURE DISTILLATION''), the derived feature aggregations are applied to perform the feature distillation between teacher and student.
Details will be described in this section.

\begin{algorithm}[tp]
    \caption{Algorithm for the two-stage DFA.}
    \label{alg}
    \begin{multicols}{2}
    \begin{algorithmic}[1]
            \Function{aggregation Search}{}
            \State Random initialize $\beta, w$
            \For{group $i = 1,2,...,G$}
                \For{iteration $k = 1,2,...,I_{s}$}
                    \State Optimize $\beta_i$ by $ \mathcal{L}_{\text{val}}(w, \beta)$
                    \State Optimize $w$ by $\mathcal{L}_{\text{train}}(w, \beta)$ 
                \EndFor
            \EndFor
            \State Reserve the derived $\beta$.
            \EndFunction
            \Function{Feature Distillation}{}
            \State Get the derived $\beta$.
            \For{iteration $k = 1,2,...,I_{v}$}
                \For{group $i = 1,2,...,G$}
                \State Calculate $A_i$ with $\beta_i$.
                \EndFor
                \State Update $w$ by minimizing the 
                \State loss defined in Eqn.~(\ref{eqn:total}).
            \EndFor
            \EndFunction
    \end{algorithmic}
    \end{multicols}
\end{algorithm}

\subsection{Feature Distillation}

DFA is based on feature distillation on $G$ \emph{layer groups}, where a layer group denotes the set of layers with the same spatial size in teacher and student networks.
The general design schemes for feature distillation are categorized into \emph{teacher transform}, \emph{student transform}, \emph{distillation position} and \emph{distance function}~\cite{Heo_2019_ICCV}. 
Teacher transform and student transform extract knowledge from hidden features of the teacher and student networks at the distillation positions respectively.
Then, the extracted features are applied to the distance function of distillation.
Most approaches~\cite{heo2019comprehensive,tung2019similarity,zagoruyko2016paying} adopt $L_2$ loss as the distance measurement. Let $N_i^T$ and $N_i^S$ denote the number of layers in the $i$-th layer group of teacher and student network, the distillation loss is defined as:
\begin{equation}
    \mathcal{L}_{\text{fd}} = \sum_{i=1}^G L_2(\mathcal{F}_t^i(T_i^{N_i^T}), \mathcal{F}_s^i(S_i^{N_i^S})))
\label{eqn:fd}
\end{equation}
where $T_i^{N_i^T}$ and $S_i^{N_i^S}$ denote the feature maps of teacher and student networks drawn from the distillation position of the $i$-th group. Conforming to the previous work~\cite{zagoruyko2016paying}, the distillation positions of the teacher and student network lay at the end of each layer group. 
Besides, $\mathcal{F}_t^i(\cdot)$ and $\mathcal{F}_s^i(\cdot)$ in Eqn.~(\ref{eqn:fd}) represent the teacher transform and student transform respectively, which map the channel numbers of both $T_i^{N_i^T}$ and $S_i^{N_i^S}$ to the channel number of teacher feature map. Traditional feature distillation methods are illustrated in Fig.~\ref{fig:compare} (a).

Different from traditional single-teacher feature distillation methods,
DFA utilizes feature aggregation of teacher network as the supervision for student network for each layer group, as shown in Fig.~\ref{fig:compare} (b).
Given the feature aggregation weights $\alpha_i = \{\alpha_i^1, ..., \alpha_i^{N_i^T}\}$ of $i$-th group in the teacher network, where $\sum_{j=1}^{N_i^T} \alpha_i^j = 1$, the feature aggregation of $i$-th group $A_i$ can be computed by:
\begin{equation}
    A_i = \sum_{j=1}^{N_i^T} \alpha_i^j \, T_i^j.
\end{equation}

The existing feature distillation methods could be seen as a special case of feature aggregation, where the weight of the last layer for each layer group $i$ of the teacher network, i.e., $\alpha_i^{N_i^T}$, is set to one and the weights of the other layers in the group are set to zero. 
Given the feature aggregation of different layer groups, the feature distillation loss in Eqn.~(\ref{eqn:fd}) is changed to:
\begin{equation}
    \mathcal{L}_{\text{fd}} = \sum_{i=1}^G L_2(\mathcal{F}_t^i(A_i), \mathcal{F}_s^i(S_i^{N_i^S}))).
\label{eqn:fd2}
\end{equation}

Finally, the student network is optimized by a weighted sum of distillation loss $\mathcal{L}_{\text{fd}}$ and classification loss $\mathcal{L}_{\text{ce}}$:
\begin{equation}
    \mathcal{L}_{\text{student}} = \mathcal{L}_{\text{ce}} + \gamma_{\text{fd}} * \mathcal{L}_{\text{fd}}
\label{eqn:total}
\end{equation}
where $\gamma_{\text{fd}}$ is the balancing hyperparameter.
$\mathcal{L}_{\text{ce}}$ is the standard cross-entropy loss between the ground-truth class label $gt$ and the output distribution of the student $p = \{p^1, ..., p^C\}$:
\begin{equation}
    \mathcal{L}_{\text{ce}}(gt, p) = -\sum_{i=1}^C \mathbb{I}[i = gt] \log(p^i),
\label{eqn:ce}
\end{equation}
where $C$ represents the number of classes and $\mathbb{I}[\cdot]$ is the indicator function.

\subsection{Differentiable Group-wise Search}\label{sec:dfa}
As the feature aggregation weights are continuous and grow exponentially with the number of layer groups, DFA leverages a differentiable architecture search method to efficiently search for the task-dependent feature aggregation weights for better distillation performance. 
Inspired by previous attempts that divide the NAS search space into blocks~\cite{liu2019darts,zoph2018learning}, DFA implements the feature aggregation search in a group-wise manner, i.e., the weights of other groups keep fixed when searching for the aggregation weights for layer group $i$. 
The group-wise search enables a strong learning capability of the model, leading to only a few epochs to achieve convergence during training. The overall framework of the differentiable group-wise search is shown in Fig.~\ref{fig:search}. 

\subsubsection{Search Space:}
Given the teacher and student networks, DFA aims to find the appropriate feature aggregation weights $\alpha_i$ for each group $i$.
Different from the DARTS-based methods~\cite{liu2019darts}, the search space for the feature aggregation is continuous since the combination of different layers could provide richer information than the individual feature map obtained from the discrete search space. 
Besides, as only one teacher network is utilized in the aggregation search, the training speed and computation overhead are similar to the standard feature distillation.
For a stable training process, we represent the feature aggregation weights $\alpha$ as a softmax over a set of architecture parameters $\beta$:
\begin{equation}
    \alpha_i^j = \frac{\exp(\beta_i^j)}{\sum_{j^{'}=1}^{N_i^T}exp(\beta_i^{j'})}.
\end{equation}

\subsubsection{Optimization of Differentiable Group-wise Search:}

The goal of the differentiable group-wise search is to \textit{jointly optimize} the architecture parameters $\beta$ and the model parameters $w$ of the student network. 
Specifically, the differentiable search tries to find the $\beta^*$ that minimizes the validation loss $\mathcal{L}_{\text{val}}(w^*(\beta), \beta)$, where the weights of the architecture parameters $w^*$ are obtained by minimizing the training loss $\mathcal{L}_{\text{train}}(w, \beta)$ for a certain architecture parameter $\beta$.
Thus, the joint optimization could be viewed as a bi-level optimization problem with $\beta$ as the upper-level variable and $w$ as the lower-level variable:
\begin{align}
    \min_{\beta} \quad  & \mathcal{L}_{\text{val}}(w^*(\beta), \beta) + \lambda \mathcal{R}(\beta) \\
    \textbf{s.t.} \quad  & w^*(\beta) = {\rm argmin}_w \mathcal{L}_{\text{train}}(w, \beta),  
\end{align}
where $\mathcal{L}_{\text{train}}$ and $\mathcal{L}_{\text{val}}$ are the training and validation loss respectively. $\mathcal{R}(\cdot)$ denotes the regularization on the architecture parameters $\beta$ that could slightly boost the performance of DFA.
To solve the bi-level optimization problem, $\beta$ and $w$ are alternately trained in a multi-step way by gradient descent to reach a fixed point of architecture parameters and model parameters.

An intuitive option of training and validation loss is to use $\mathcal{L}_{\text{student}}$ in Eqn.~(\ref{eqn:total}). 
Though it seems that directly learning from $\mathcal{L}_{\text{student}}$ could result in the appropriate architecture parameters for knowledge distillation, actually, training architecture parameters with $\mathcal{L}_{\text{student}}$ is equivalent to minimizing the distillation loss between feature aggregation and student feature map:
\begin{equation}
    \arg\min_{\beta} \,  \mathcal{L}_{\text{student}}(w^*(\beta), \beta) = \arg\min_{\beta} \,  \mathcal{L}_{\text{fd}}(w^*(\beta), \beta),
\end{equation}
as the cross-entropy loss $\mathcal{L}_{\text{ce}}$ is irrelevant to the architecture parameters.
Since the distillation loss only characterizes the distance between the student feature maps and the combinations of the teacher feature maps, the architecture parameters tend to be more inclined to choose teachers that are close to the student.
In depth, suppose that after training several epochs, the student feature map has learnt some knowledge from the data distribution through cross-entropy loss and teacher network through distillation loss.
As the student network is always shallower than the teacher network,
the knowledge in the deep layers is hard to learn such that
the architecture parameters would prefer the teachers in the shallow layers matching the depth and expressivity of the student, other than selecting deep layers with rich semantics and strong expressivity.
Therefore, the feature aggregation learnt from the $\mathcal{L}_{\text{student}}$ deviates from the original target that learning a good teacher for the knowledge distillation.
Besides, once the student network finds a matching layer in the teacher group, i.e., the weight of an architecture parameter $\beta_i^j$ is relative larger than the others, the student transform will learn more about the mapping function from $S_i^{N_i^S}$ to $T_i^j$.
Then, $\beta_i^j$ will grow much faster than other competitors due to the biased training of transform function, and the corresponding feature map will gradually dominant the feature aggregation under the exclusive competition of the architecture parameters.
Notice again that the network is more likely to pick shallow layers in the early training stage.
Therefore, the student network would unavoidably suffer from the performance collapse using the search results derived from $\mathcal{L}_{\text{student}}$.

\begin{figure*}[t]
\centering
\includegraphics[width=0.8\textwidth]{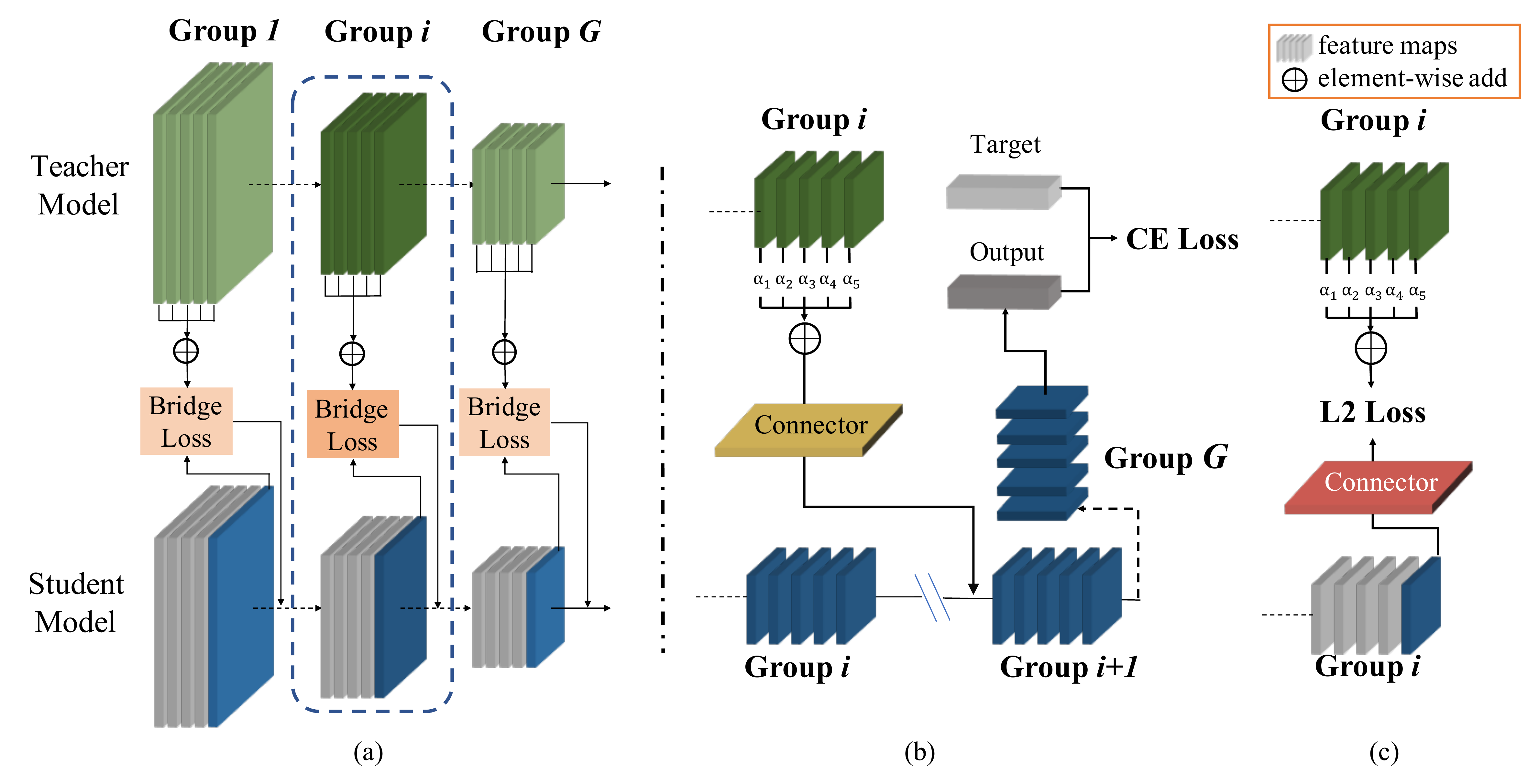}
\caption{Differentiable group-wise search of DFA. (a) The differentiable search for group $i$. (b) The teacher-to-student (TS) path. (c) The student-to-teacher (ST) path. The ST and TS connectors are implemented with $1 \times 1$ convolutional layers for matching the channel dimensions between teacher and student.}
\label{fig:search}
\end{figure*}

\subsubsection{Bridge Loss for Feature Aggregation Search:}

To search for an appropriate feature aggregation for the knowledge distillation, we introduce the \emph{bridge loss} to connect the teacher and student networks, where the original information flow of the student network is split into two paths.

In the first teacher-to-student (TS) path as illustrated in Fig.~\ref{fig:search} (b), DFA takes the feature aggregation of the group $i$ in the teacher network, i.e., $A_i$, as the input of its $(i+1)$-th group of the student network, and then computes the teacher-to-student (TS) loss $\mathcal{L}_{\text{TS}}$ with standard cross entropy. The TS loss in group $i$ can be expressed as:

\begin{equation}
    \mathcal{L}_{\text{TS}}^{i} = \mathcal{L}_{\text{ce}}(gt, f_G(...(f_{i+1}(\mathcal{F}_t^i(A_i)))))
\end{equation}
where $gt$ denotes the ground-truth class label and $f_i$ denotes the convolutional layers of group $i$ in the student network.
Different from Eqn.~(\ref{eqn:fd2}), the teacher transform $\mathcal{F}_t^i$ is now served as a TS connector that converts the channel dimension of the teacher feature map to the student feature map.

The second student-to-teacher (ST) path has the similar effect as the original $\mathcal{L}_{\text{student}}$ in Eqn.~(\ref{eqn:total}), i.e., exploring the feature aggregation weights that match the learning ability of the student network. As shown in Fig.~\ref{fig:search} (c), the information flow starts from the student input and ends at the $i$-th group in the student network. Then, the student network produces $S_i^{N_i^S}$, the last feature map of group $i$, and compares $S_i^{N_i^S}$ with the feature aggregation $A_i$ by a student-to-teacher (ST) loss $\mathcal{L}_{\text{ST}}$:
\begin{equation}
    \mathcal{L}_{\text{ST}}^i = L_2(\frac{\vectorize{\mathcal{F}_s^i(S_i^{N_i^S})}}{||\vectorize{\mathcal{F}_s^i(S_i^{N_i^S})}||_2}, \frac{\vectorize{A_i}}{||\vectorize{A_i}||_2})
\end{equation}
where $\vectorize{\cdot}$ represents the vectorization of the tensor that converts the tensor into a column vector.
Same as the distillation loss in Eqn.~(\ref{eqn:fd2}), the student transform $\mathcal{F}_s^i$ is served as a ST connector to map the channel numbers from student feature map to teacher feature map, as opposed to $\mathcal{F}_t^i$.

For each group $i$, the bridge loss $\mathcal{L}_{\text{Bridge}}^i$ integrates ST loss $\mathcal{L}_{\text{ST}}$ and TS loss $\mathcal{L}_{\text{TS}}$ in a single training process for both training loss $\mathcal{L}_{\text{train}}$ and validation loss $\mathcal{L}_{\text{val}}$:
\begin{equation}\label{eqn:bridge}
    \mathcal{L}_{\text{Bridge}}^i = \gamma_{\text{ST}}\mathcal{L}_{\text{ST}}^i + \gamma_{\text{TS}}\mathcal{L}_{\text{TS}}^i.
\end{equation}
where $\gamma_{\text{ST}}$ and $\gamma_{\text{TS}}$ are balancing hyperparameters. 

Different from $\mathcal{L}_{\text{student}}$, both the model parameters and architecture parameters can be trained towards the ground truth through bridge loss.
For the model parameters, the student network before $(i+1)$-th group tries to imitate the teacher feature aggregation $A_i$, while the student after $(i+1)$-th group learns to optimize the cross entropy given $A_i$.
Hence, the joint optimization of ST loss and TS loss for the model weights can be regarded as an approximation of the cross entropy loss of the student network.
On the other hand, the architecture parameters are trained by directly optimizing the cross-entropy loss $\mathcal{L}_{\text{TS}}$ given the input of the aggregation of teacher feature maps, analogous to the common differentiable architecture search in~\cite{liu2019darts,xie2019snas}.
The deep layers with rich features will be assigned by higher weights in the architecture search, as they contribute to the reduction of the validation loss.
In this case, DFA achieves the feature aggregation which helps the student learn a wealth of knowledge from the teacher.
Besides, we still keep ST loss in the architecture training as a regularization.
Specifically, though the rich features in the deep layers contribute better performance in the teacher network, they are not always suitable for the student to learn due to the mismatch of expressive power, e.g., it is impractical for a student of three layers to learn from the teacher of ten layers.
Introducing ST loss in the validation loss can help the network select the shallow teachers that match the student expressivity, such that the derived feature could provide more knowledge to the student in the early training stage and accelerate the model convergence.
In this way, the ST loss and TS loss act as two players against each other, trying to optimize the unified architecture parameters to the opposite directions while guaranteeing both expressivity and learnability of the feature aggregation simultaneously. 
Hence, the derived feature 
is more likely to achieve better performance in the final knowledge distillation.


After searching for the architecture parameters with the differentiable group-wise search, DFA trains the student network thoroughly with the derived feature aggregation weights by Eqn.~(\ref{eqn:total}).

\subsection{Time Complexity Analysis}

The two-stage design of DFA would not increase the time complexity compared with other feature distillation methods. Let $t_T^i, t_S^i$ denote the computing time of layer group $i$ in the teacher and student network, the differentiable search in the first stage could be calculated by:


\begin{equation}
\mathcal{T}_1 = (\sum_{j=1}^{i}(t_T^j + t_S^j)) + (\sum_{j=1}^it_T^j + \sum_{j=i+1}^Gt_S^j) = 2\sum_{j=1}^it_T^j + \sum_{j=1}^Gt_S^j.
\end{equation}
As the second stage is a usual feature distillation, the overall time complexity of DFA could be derived as:
\begin{equation}
\begin{aligned}
    \mathcal{T} &= \mathcal{T}_1 + (\sum_{j=1}^Gt_T^j + \sum_{j=1}^Gt_S^j)= 2\sum_{j=1}^it_T^j + \sum_{j=1}^Gt_T^j + 2\sum_{j=1}^Gt_S^j \\  
    &= O(\sum_{j=1}^Gt_T^j) + O(\sum_{j=1}^Gt_S^j)
\end{aligned}
\end{equation}
which is competitive with other feature distillation methods.

\subsection{Implementation Details}
The ST connector (student transform) and TS connector (teacher transform) are both implemented with the one-layer convolution networks in order to reconcile the channel dimensions between student and teacher feature maps. 
The weight of ST loss and TS loss are set to $\gamma_{\text{ST}}=1e-3$ and $\gamma_{\text{TS}}=1$ in Eqn.~(\ref{eqn:bridge}).
In both stage of DFA, the pre-ReLU features are extracted in the student and teacher networks for the knowledge distillation, where values no smaller than -1 are preserved in the feature maps to retain knowledge from both positive values and negative values while avoiding the exploding gradient problem. The model parameters are initialized by He initialization~\cite{he2015delving}. The feature aggregation weights in each layer group are initialized in the way that only the last feature map has weight $1$ and all other feature maps are allocated zero weight.

We follow the same training scheme as DARTS: only the training set are used to update model parameters, and the validation set are leveraged for better feature aggregation parameters.
For the update of the architecture parameters, DFA adopts Adam~\cite{kingma2014adam} as the optimizer with the momentum of $(0.5,0.999)$, where the learning rate and weight decay rate are both set to 1e-3.

\section{Experiments}

\subsection{CIFAR-100}
\label{exp:cifar100}
CIFAR-100 is a commonly used visual recognition dataset for comparing distillation methods. There are 100 classes in CIFAR-100, and each class contains 500 training images and 100 testing images. 
To carry out architecture search,
the original training images are divided into the training set and validation set with the 7:3 ratio. 

We compare our method with eight single-teacher distillation methods: \textit{KD}~\cite{hinton2015distilling}, \textit{FitNets}~\cite{romero2014fitnets}, \textit{AT}~\cite{zagoruyko2016paying}, \textit{Jacobian}~\cite{srinivas2018knowledge}, \textit{FT}~\cite{kim2018paraphrasing}, \textit{AB}~\cite{heo2019knowledge}, \textit{SP}~\cite{tung2019similarity} and \textit{Margin}~\cite{heo2019comprehensive}. 
All the experiments are performed on Wide Residual Network~\cite{zagoruyko2016wide}. 
The feature aggregation is searched for 40 epochs at each layer group.
We adopt the same training schemes as~\cite{zagoruyko2016wide} to train the model parameters in all methods. 
Specifically, the model is trained by SGD optimizer of 5e-4 weight decay and 0.9 momentum for 200 epochs on both training and validation set.
The learning rate is set to 0.1 initially and decayed by 0.2 at 60, 120 and 160 epochs. 
The batch size is 128. 
We utilize random crop and random horizontal flip as the data augmentation.



\begin{wraptable}[10]{r}{0.49\textwidth}
\centering
\begin{tabular}{c|cc|cc}
\hline
    & Teacher  & Size & Student  & Size \\
\hline
(1) & WRN\_28\_4 & 5.87M  & WRN\_16\_4 & 2.77M \\
(2) & WRN\_28\_4 & 5.87M  & WRN\_28\_2 & 1.47M \\
(3) & WRN\_28\_4 & 5.87M  & WRN\_16\_2 & 0.7M  \\
\hline
\end{tabular}
\caption{The configuration of teacher and student networks in experiments on CIFAR-100.}
\label{tab:cifar100_teacher_student}
\end{wraptable}

We explore the performance of our method on several teacher-student pairs, which vary in depth (number of layers), width (number of channels), or both, as shown in Table~\ref{tab:cifar100_teacher_student}.
We conduct the experiments on CIFAR-100 over the above teacher-student pairs, and depict the results in Table~\ref{tab:cifar100}.
For the teacher-student pair (1) of different widths, DFA has a 1.34\% improvement over the output distillation method KD, and also outperforms the other state-of-the-art feature distillation methods. DFA even exhibits a better performance than the teacher network. 
For the teacher-student pair (2) of different depths, DFA surpasses other feature distillation methods by 0.34\%-2.64\%.
DFA also achieves state-of-the-art results on (3), where the student network compresses both width and depth of the teacher network. 
The above experiments verify the effectiveness and robustness of DFA in various scenarios.

\begin{table*}[htbp]
\centering
\begin{tabular}{c|cc|cccccccc|c}
\hline 
   & Teacher & Student & KD   & FitNets &  AT  & Jacobian &   FT   &  AB   &   SP  & Margin & DFA \\
\hline 
(1)& 79.17   &  77.24  & 78.4  & 78.17   & 77.49 &  77.84   & 78.26 & 78.65 & 78.7  & 79.11  & \textbf{79.74} \\
(2)& 79.17   &  75.78  & 76.61 & 76.14   & 75.54 &  76.24   & 76.51 & 76.81 & 77.41 & 77.84  & \textbf{78.18} \\ 
(3)& 79.17   &  73.42  & 73.35 & 73.65   & 73.3  &  73.28   & 74.17 & 73.9  & 74.09 & 75.51  & \textbf{75.85} \\
\hline
\end{tabular}
\caption{The experiment results on CIFAR-100. We compare the proposed DFA with eight distillation methods. The best results are illustrated in bold. DFA outperforms other state-of-the-art methods.}
\label{tab:cifar100}
\end{table*}

\subsection{CINIC-10}
CINIC-10 is a large classification dataset containing 270000 images, which are equally split into training, validation and test set with the presence of 10 object categories. The images are collected from CIFAR-10 and ImageNet. Comparing with CIFAR datasets, CINIC-10 could present a more principled perspective of generalisation performance. We explore the performance of DFA and other three state-of-the-art feature distillation methods on CINIC-10. All the experiments are performed on ShuffleNetV2~\cite{ma2018shufflenet}, an efficient network architecture at mobile platforms. We use several variants of ShuffleNetV2 in the experiments, and the basic configuration is shown in Table~\ref{tab:shufflenet} conforming to~\cite{tung2019similarity}. 
In the model training process, the SGD optimizer is leveraged, with weight decay of 5e-4 and momentum of 0.9. All models are trained with 140 epochs. The learning rate is set to 0.01 initially and decayed by 0.1 at 100, 120 epochs. The batch size is 96. Same as CIFAR dataset, we utilize random crop and random horizontal flip as the data augmentation. We search 5 epochs for each feature group in DFA. The experiment results are shown in Table~\ref{tab:cinic10}. DFA outperforms the vanilla cross-entropy training as well as the state-of-the-art feature distillation methods.

\begin{table}
\centering
\resizebox{0.6\textwidth}{!}{
\begin{tabular}{c|c|ccc|c}
\hline 
Group & Block & $k$ & $c$ & $n$ & Output Size \\
\hline 
1 & Conv-BN-ReLU & 3 & 24 & 1 & $32\times32$ \\
2 & ShuffleNetV2 block & 3 & $116x$ & 4 & $16\times16$ \\
3 & ShuffleNetV2 block & 3 & $232x$ & 8 & $8\times8$ \\
4 & ShuffleNetV2 block & 3 & $464x$ & 4 & $4\times4$ \\
5 & ShuffleNetV2 block & 3 & $1024*max(1,x)$ & 1 & $4\times4$ \\
\hline
6 & AvgPool-FC & 1 & 10 & 1 & $1\times1$ \\
\hline
\end{tabular}
}
\caption{The configuration of ShuffleNetV2 on CINIC-10 experiments. We leverage standard ShuffleNetV2 blocks in each layer group, where $k$ denotes the kernel size, $c$ and $n$ specify the number of channels and blocks in each layer group. In the end, we add an average pooling layer and a fully connected layer to make final predictions.}
\label{tab:shufflenet}
\end{table}

\begin{table}
\centering
\resizebox{0.8\textwidth}{!}{
\begin{tabular}{c|ccc|ccc|ccc|cc}
\hline 
& Teacher & Size & Acc & Student & Size & Acc & AT &   SP  & Margin & DFA & DFA-T\\
\hline 
(1) & $x=2.0$ & 5.37M & 86.14 & $x=1.0$ & 1.27M & 83.28 & 84.71 & 85.32 & 85.29 & 85.38 & \textbf{85.41} \\
(2) & $x=2.0$ & 5.37M & 86.14 & $x=0.5$ & 0.36M & 77.34 & 79.06 & 79.15 & 78.79 & \textbf{79.51} & 79.45 \\
(3) & $x=1.0$ & 1.27M & 83.28 & $x=0.5$ & 0.36M & 77.34 & 78.42 & 79.02 & 79.69 & \textbf{79.97} & 79.38 \\
\hline
\end{tabular}
}
\caption{Experimental results on CINIC-10. The best results are illustrated in bold. ``DFA-T'' represents the version that searches the feature aggregation weights on CIFAR-100 and implement the feature distillation on CINIC-10.}
\label{tab:cinic10}
\end{table}

\subsubsection{Ablation Study on Differentiable Search:}

We study the robustness of DFA by searching the feature aggregation weights from a small dataset and then distillating on a larger dataset. Specifically, we build up a variant of DFA, named as DFA-T, by searching the feature aggregation weights on CIFAR-100 and distillating on CINIC-10. 
It can be observed that DFA-T is only little inferior to DFA, and still achieves state-of-the-art performance.

\subsection{The Effectiveness of Differentiable Search}

\subsubsection{Comparisons with Other Search Methods:}

We perform additional experiments on CIFAR-100 to verify the effectiveness of the differentiable feature aggregation search. We compare DFA with the following methods: "\textit{Random}" represents the method that all feature aggregation weights are randomly selected; "\textit{Average}" indicates that all feature maps in a layer group share the same feature aggregation weight; "\textit{Last}" denotes the method that only the last feature map in each layer group is leveraged for the feature distillation, which is widely used in the knowledge distillation. The results are shown in Table~\ref{tab:ablation}. 

\begin{wraptable}[11]{r}{0.49\textwidth}
\centering
\resizebox{0.5\textwidth}{!}{
\begin{tabular}{c|c|c|c}
\hline 
Method  & WRN\_16\_2 & WRN\_16\_4 & WRN\_28\_2 \\
\hline
Student & 73.42      & 77.24      & 75.78      \\
Random  & 74.25      & 78.92      & 77.07      \\
Last  & 75.51      & 79.11      & 77.86      \\
Average & 74.11      & 78.81      & 76.99      \\
\hline
DFA     & \textbf{75.85} & \textbf{79.74} & \textbf{78.18} \\
\hline 
\end{tabular}
}
\caption{Experimental results of DFA and other search methods on CIFAR-100.}
\label{tab:ablation}
\end{wraptable}

Obviously, ``Random'' weights or ``Average'' weights would degrade the performance, indicating the necessity of a decently designed feature selection strategy.
Different from ``Last'', we observe that DFA would allocate positive weights to the shallow layers in the shallow groups to retrieve knowledge rapidly from the teacher network.
The weight assignment in DFA reveals that feature aggregation contributes to transferring knowledge from the teacher network to the student network, while the differentiable group-wise search helps achieve the optimal feature aggregation weights.
Hence, DFA brings about a remarkable improvement on the knowledge distillation task.

\subsubsection{Result Analysis:}
In Fig.~\ref{fig:heatmap&sen} \textit{Left}, we display the student network's feature aggregation weights in the configuration (1) of the CIFAR-100 experiment. The feature aggregation weights are initialized with the ``Last'' scheme and then searched for $40$ epochs. It is obvious that the domination of the last feature map in each layer group are weakened as the training continues.  

\subsubsection{Sensitivity Analysis:}
\begin{figure*}
\begin{minipage}[t]{0.5\linewidth}
   \includegraphics[width=\linewidth]{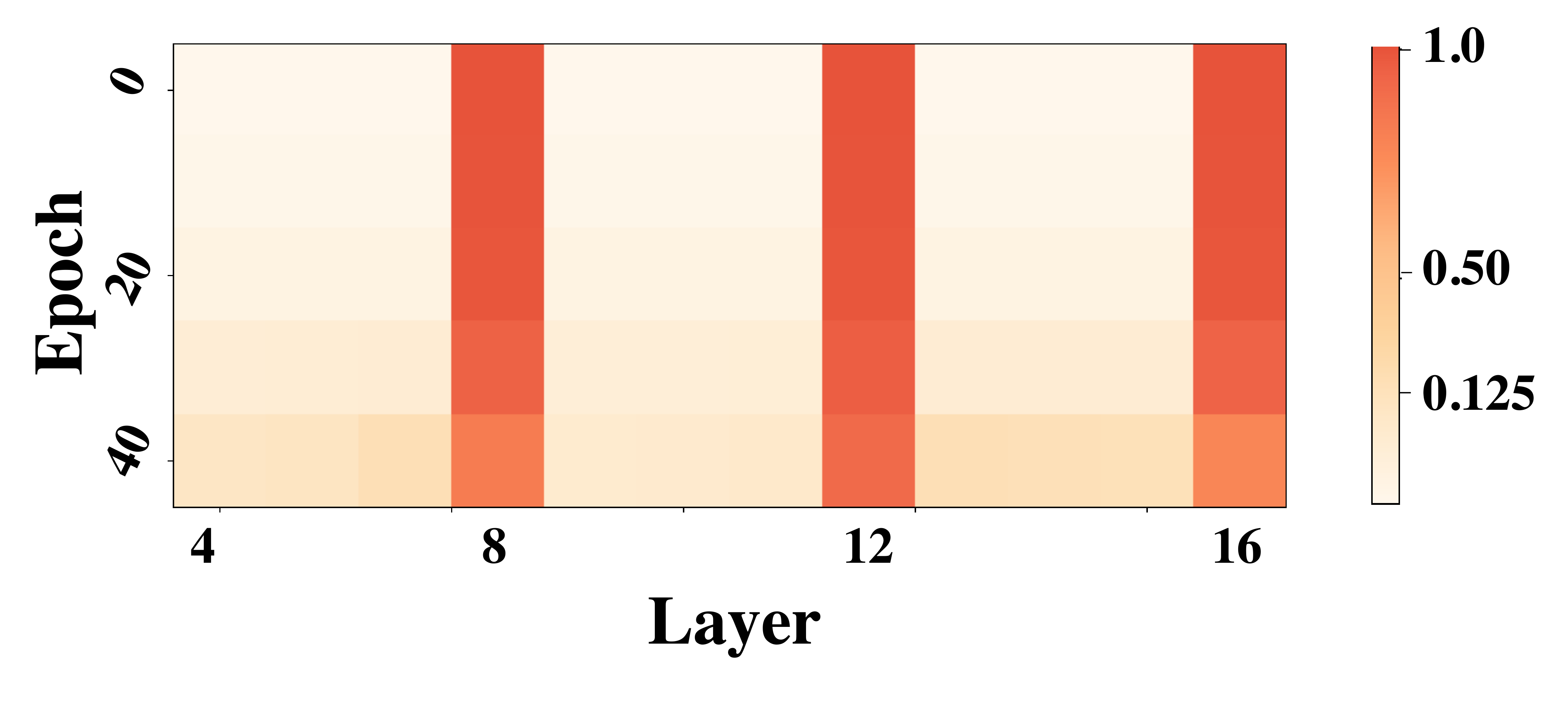}
  \end{minipage}
  \begin{minipage}[t]{0.5\linewidth}
    \includegraphics[width=\linewidth]{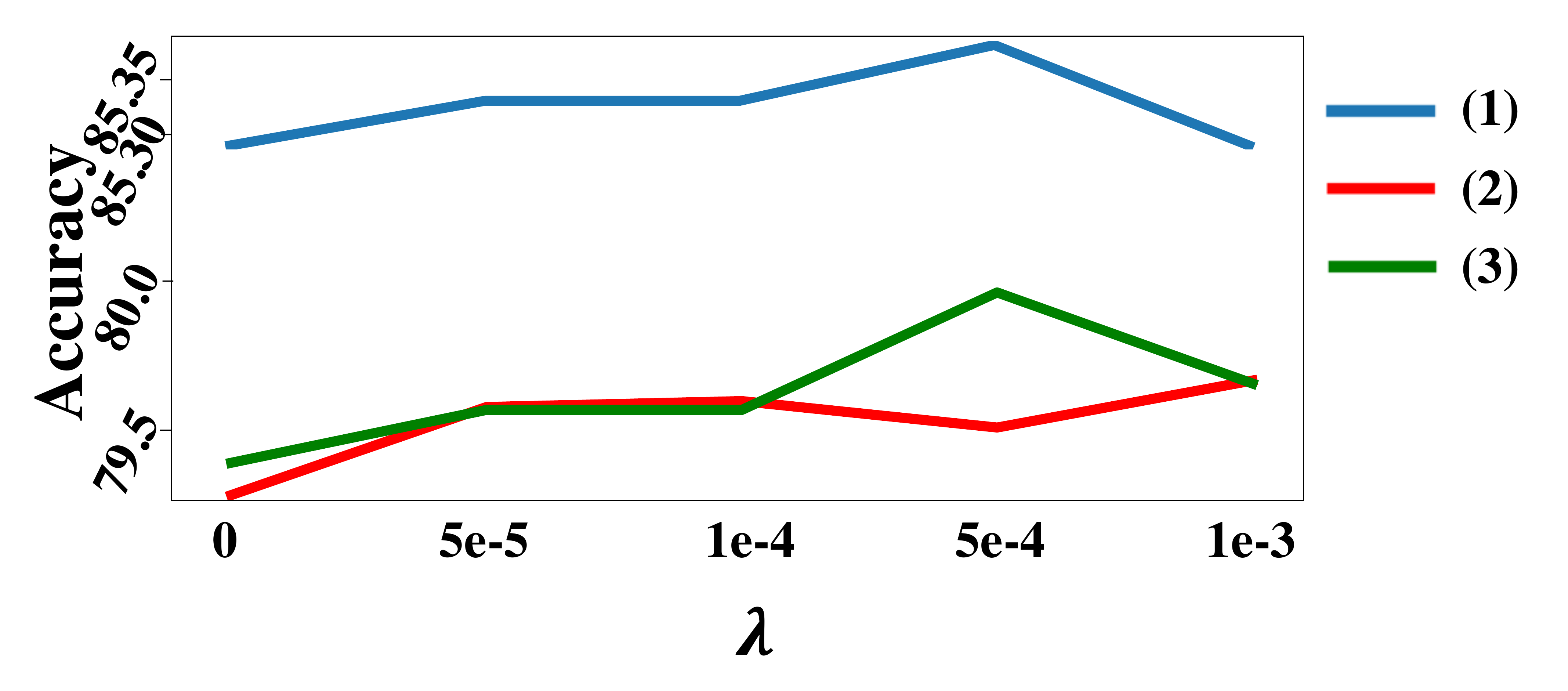}
  \end{minipage}
  \caption{\textit{Left.} Heatmap of feature aggregation weights in different layers of WRN\_16\_4. \textit{Right.} The accuracy of the student network on CINIC-10, $\lambda$ ranges from $[0,1e-3]$. We select the three teacher and student pairs in the CINIC-10 experiments.}
  \label{fig:heatmap&sen}
\end{figure*}

We study the impact of regularization on the feature aggregation search mentioned in Sec.~\ref{sec:dfa}. The right figure in Fig.~\ref{fig:heatmap&sen} displays the accuracy of the student models with $\lambda$ ranging from $0$ (no regularization) to $1e-3$. Experimental results show that DFA is robust to the regularization in range $(0, 1e-3]$, except a slight decrease without regularization.

\section{Conclusion}
In this paper, we propose DFA, a two-stage feature distillation method via differentiable aggregation search. In the first stage, DFA leverages the differentiable architecture search to find appropriate feature aggregation weights. It introduces a bridge loss to connect the teacher and student, where a teacher-to-student loss is built for searching the teacher with rich features and a wealth of knowledge, while a student-to-teacher loss is used to find the aggregation weights that match the learning ability of the student network. In the second stage, DFA performs a standard feature distillation with the derived feature aggregation weights. 
Experiments show that DFA outperforms several state-of-the-art methods on CIFAR-100 and large-scale CINIC-10 datasets, verifying both the effectiveness and robustness of the design. 
In-depth analysis also reveals that DFA decently allocates feature aggregation weights on the knowledge distillation task.

\section*{Acknowledgment}
This work is partially supported by National Key Research and Development Program No. 2017YFB0803302, Beijing Academy of Artificial Intelligence (BAAI), and NSFC 61632017.

\clearpage
%
%
\bibliographystyle{splncs04}
\bibliography{eccv20}
\end{document}